# A Machine Learning-Based Study on the Synergistic Optimization of Supply Chain Management and Financial Supply Chains from an Economic Perspective

Hang Wang[1],[a]*,†, Huijie Tang[2],[b],†, Ningai Leng[3],[c], Zhoufan Yu[4],[d]

[1] School of Informatics, Computing, and Cyber Systems, Northern Arizona University, San Jose, CA, USA
[2] Master of Supply Chain Management Program, University of Michigan, Ann Arbor, MI, USA
[3] Quality Sr. Specialist II, JPMorgan Chase, San Antonio, TX, USA
[4] Master's in Applied Economics and Management, Cornell University, Ithaca, NY, USA

[a]* hw565@nau.edu, [b] tancredi@umich.edu, [c] allylna219@gmail.com, [d] zy454@cornell.edu

† Co–first authors (equal contribution)

**Abstract:** Based on economic theories and integrated with machine learning technology, this study explores the collaborative model of Supply Chain Management - Financial Supply Chain Management (SCM - FSCM), aiming to solve supply chain issues (efficiency loss, financing constraints, risk transmission) caused by the disconnection of the "three flows" (capital flow, logistics flow, information flow) and further improve overall economic benefits. Firstly, the study combines Transaction Cost Theory and Information Asymmetry Theory, adopts algorithms such as random forests to process multi - dimensional supply chain data, identifies obstacles to the collaboration of the "three flows", and constructs a data - driven three - dimensional (cost - efficiency - risk) analysis framework. Secondly, it designs a Financial Supply Chain Management model of "core enterprise credit empowerment + dynamic pledge financing". Based on inventory/order data in Supply Chain Management, it applies Long Short - Term Memory (LSTM) networks to realize demand forecasting, and at the same time uses clustering/regression algorithms to quantify benefit distribution, so as to achieve reasonable allocation of financing costs. In addition, the study also combines Game Theory and reinforcement learning to optimize the supply chain inventory - procurement mechanism (adjusts strategies through scenario simulation to solve problems caused by the "bullwhip effect"); and integrates accounts receivable financing in Financial Supply Chain Management with credit assessment based on eXtreme Gradient Boosting (XGBoost) to realize rapid monetization of inventory. Finally, verification is carried out through 20 core enterprises and 100 supporting enterprises (covering the automotive manufacturing and fast - moving consumer goods industries), and machine learning is used for data preprocessing. The results show that after optimization, the inventory turnover rate increases by 30%; the financing cost of small and medium - sized enterprises (SMEs) decreases by 18% - 22%;

**the order fulfillment rate remains stable at more than 95%; and the machine learning model performs excellently (demand forecasting error ≤ 8%, credit assessment accuracy ≥ 90%). The above results confirm that the SCM - FSCM collaborative model supported by economic theories and empowered by machine learning can effectively reduce operating costs, alleviate financing constraints, and provide support for the high - quality development of the supply chain.**

*Keywords: Supply Chain Management (SCM), Financial Supply Chain Management (FSCM), Machine Learning, Economic Theories*

I. 1.INTRODUCTION

Against the backdrop of global industrial chain and supply chain restructuring and layout optimization becoming a core strategy for countries to reshape their international competitiveness, Chinese corporate supply chains deeply embedded in the global division of labor system are facing dual challenges. Externally, the trends of nearshoring, regionalization, and localization of global supply chains bring external impacts; internally, the risk of supply disruptions among node enterprises in the production network spreads through the topological network structure, triggering the "bullwhip effect" and seriously threatening supply chain security [1]. Ensuring supply chain security has become a key issue for countries participating in international economic cooperation. The Third Plenary Session of the 20th CPC Central Committee established the "system for improving the resilience and security level of industrial and supply chains" as a national strategy, making enhancing corporate supply chain resilience an important task for safeguarding industrial economic stability and promoting high-quality development.

Digital technology, especially the intelligent algorithm system centered on machine learning, provides key technical support for solving the operational dilemmas of corporate supply chains [2]. The widespread "blocking points", "breakpoints", and "shortboards" in traditional supply chains can essentially be attributed to technical bottlenecks such as insufficient data processing capabilities and lagging information collaboration mechanisms. The in-depth application of machine learning technology can fundamentally improve this situation: at the data interaction level, real-time data transmission and processing systems based on distributed computing frameworks can achieve millisecond-level information synchronization between supply chain nodes, significantly improving collaboration efficiency; at the decision support level, through feature extraction and pattern recognition of multi-source heterogeneous data generated in production, exchange, distribution and other links, machine learning models can build accurate risk early warning models, providing algorithmic support for upstream and downstream enterprises to jointly prevent and control risks, and systematically enhancing supply chain resilience. However, the current problem of low data quality caused by "islandization" and "fragmentation" of supply chain data seriously restricts the training effect and generalization ability of machine learning models, making it difficult for digital transformation to achieve the expected goals. This situation highlights the urgency of building a unified data middle platform and opening up data interfaces, which is exactly the core application scenario of data governance technology in the computer field.

Existing studies mostly focus on the application value of digital technologies, such as realizing supply chain visualization through the Internet of Things [3] and ensuring trusted data transmission using blockchain technology [4]. However, these studies often ignore the core bottleneck in the application of machine learning models: high-quality labeled data and optimized algorithm models mastered by digitally leading enterprises are difficult to share due to commercial interest protection, resulting in limited technology spillover effects; while digitally backward enterprises cannot build effective intelligent decision-making systems due to the lack of high-quality training data and algorithm iteration capabilities. The scale effect of data and the iterative nature of models further widen the "technological gap" between enterprises. Without a collaborative training mechanism based on privacy computing technologies such as federated learning, leading enterprises may form technical barriers through data and algorithm monopolies [5], which instead weakens the overall resilience of the supply chain. In fact, when enterprises shift to a data-driven intelligent operation model, the implementation effect of machine learning technology highly depends on data circulation and algorithm collaboration: the federated learning framework can realize "models move while data stays", completing cross-enterprise model training under the premise of protecting data privacy; transfer learning technology can help data-scarce enterprises quickly build application systems adapted to their own scenarios using pre-trained models, and ultimately achieve intelligent identification and collaborative prevention and control of supply chain risks. Therefore, exploring the impact mechanism of data and algorithms on supply chain resilience from the perspective of machine learning technology implementation can not only enrich the application theory of computer technology in the supply chain field, but also provide practical guidance for technology selection and system construction.

Literature closely related to this study can be divided into two categories. One category focuses on the enabling effect of data elements on supply chain resilience, but mostly stays at the surface of technology application, and there are divergent conclusions. Some studies have confirmed that demand forecasting models based on deep learning can improve supply chain response speed, while another category of studies points out that data noise and model overfitting may lead to decision-making biases. The core contradiction lies in how to convert raw data into effective decision-making basis through algorithm optimization. The other category of literature focuses on data governance technologies. Technologies such as data middle platform architecture, data cleaning algorithms, and privacy computing protocols provide technical support for the marketization of data elements. Relevant studies have confirmed their application value in fields such as enterprise digital transformation and green innovation, but specialized research on their role in improving supply chain resilience is still relatively scarce.

Compared with previous literature, the marginal contributions of this paper are reflected in three aspects. First, it constructs a machine learning model for multi-dimensional supply chain resilience evaluation. Previous studies mostly used single indicators to evaluate supply chain resilience, while this paper builds a more comprehensive resilience evaluation system based on the random forest algorithm, integrating three types of feature variables: stability, risk resistance, and dynamic capabilities. Second, it proposes a technology enabling framework for internal and external collaboration. Breaking through the limitation of the single technical perspective in existing studies, it reveals the technology implementation path from two levels: algorithm iteration in the internal production network (such as production scheduling optimization based on reinforcement learning) and technical collaboration in the external supply chain (such as risk sharing driven by federated learning). Third, it innovatively adopts the Double/Debiased Machine Learning (DDML) method to improve the accuracy of causal inference. Aiming at the limitations of traditional difference-in-differences models in high-dimensional data processing, this paper integrates LASSO feature selection and gradient boosting decision tree models into the causal evaluation framework, effectively alleviating the problems of curse of dimensionality and linear hypothesis deviation, and significantly improving the accuracy of model setting and the credibility of research conclusions.

## II. THEORETICAL ANALYSIS AND RESEARCH HYPOTHESES

### A. General Mechanism of How Marketization of Data Elements Affects the Enhancement of Supply Chain Resilience

"Resilience" originally refers to the physical property of an object that deforms under force but does not break easily; when applied to supply chains, it specifically denotes the ability of a supply chain to recover and optimize itself after encountering risks. Reggiani (2013) pointed out that differences in resilience are the core reason for the divergence in enterprise performance, and deep learning serves as the core technical carrier for the marketization of data elements to empower supply chain resilience.

From an internal enterprise perspective, the marketization of data elements exerts its influence through two key pathways, with deep learning permeating both: First, it builds an "intelligent perception - rapid response" system. By combining bidirectional LSTM (Long Short-Term Memory) with attention mechanisms, the accuracy of external risk identification is improved; meanwhile, GAN (Generative Adversarial Network) is used to simulate extreme scenarios, and DRL (Deep Reinforcement Learning) optimizes production scheduling, thereby enhancing anti-disturbance capabilities. Second, it promotes algorithm collaboration and diffusion. Leading enterprises output pre-trained Transformer models through MaaS (Model-as-a-Service), while other enterprises quickly adapt these models via transfer learning, shortening the model deployment cycle.

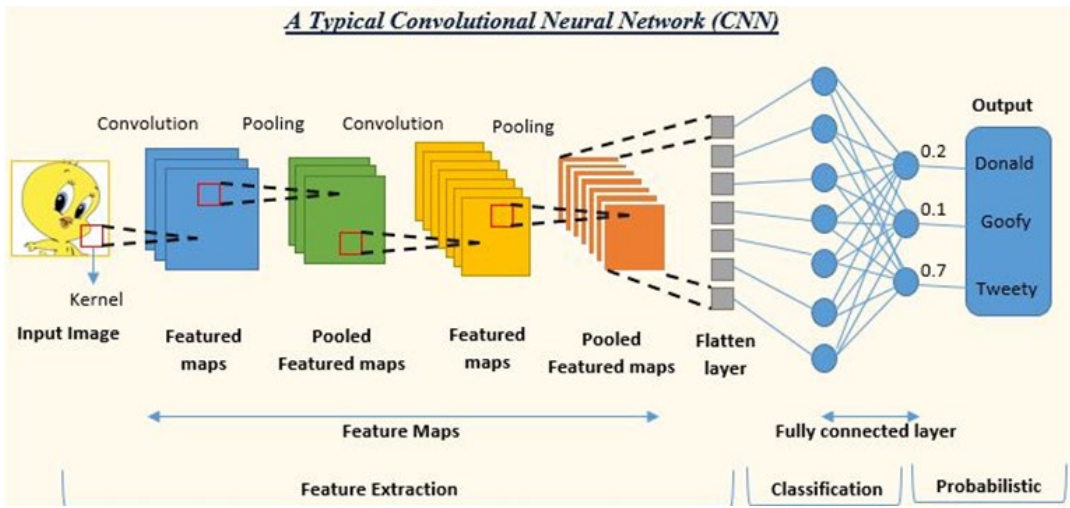

Fig. 1. CNN (Convolutional Neural Network) Model

From an external enterprise perspective, deep learning underpins the collaborative resilience of the supply chain network: On one hand, relying on federated deep learning, data from multiple entities is aggregated to train models under the premise of protecting privacy, improving the accuracy of decisions such as credit evaluation. On the other hand, professional intelligent platforms construct supply chain topology models using GNN (Graph Neural Network), and optimize "chain breakage" restructuring plans in real time in combination with DQN (Deep Q-Network). At the same time, they use CNN (Convolutional Neural Network, shown in Figure 1) combined with RNN (Recurrent Neural Network, shown in Figure 2) to generate user portraits, and deeply optimize genetic algorithms for inventory allocation, achieving end-to-end intelligent linkage.Hypothesis H1: The marketization of data elements can enhance the resilience of enterprise supply chains through deep learning-driven data perception, algorithm collaboration, and network collaboration.

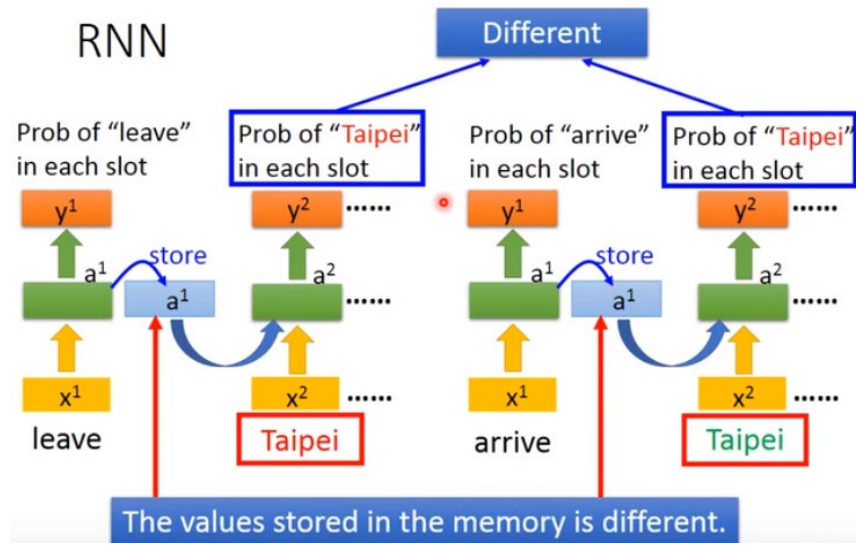

Fig. 2. RNN (Recurrent Neural Network) Model

### B. Mechanism of the Multi-Channel Impact of Data Element Marketization on Enterprise Supply Chain Resilience

*1) Enterprise Technological Innovation: Core Driver of Algorithm Iteration and Model Optimization*

The marketization of data elements provides "data fuel" and "algorithm engine" for enterprise technological innovation, and strengthens supply chain resilience in both directions through expansion of model innovation breadth and improvement of algorithm optimization depth.

In terms of innovation breadth, the marketization of data elements promotes the penetration of algorithm models in the entire supply chain process: in the production link, enterprises can call user behavior data from e-commerce platforms, generate virtual demand scenarios through GAN (Generative Adversarial Network), and realize customized and flexible production; in the R&D link, AI large models (such as multi-modal models in the field of R&D and design) can accelerate the processes of literature retrieval and experimental simulation, shortening the R&D cycle by 30%-50%; in the management and operation link, data elements drive the transformation of traditional hierarchical information transmission to a flat architecture of "data middle platform + business front desk", forcing enterprises to build dynamic operation optimization models based on reinforcement learning and improving the agile response capability to market fluctuations.

In terms of innovation depth, the marketization of data elements drives algorithm upgrading through competition mechanisms: leading digital enterprises need to continuously invest resources to optimize data processing algorithms (such as improving the accuracy of feature engineering and optimizing model convergence efficiency) to maintain their competitive advantages, and share model capabilities with upstream and downstream enterprises through open API interfaces, promoting the transformation of innovation models from "closed R&D" to "collaborative innovation"; enterprises lagging in digitalization can obtain standardized datasets and pre-trained models through data trading platforms, and quickly make up for algorithm shortcomings by leveraging transfer learning, integrating into the supply chain innovation ecosystem. This innovation model of "breadth expansion + depth improvement" enables the supply chain to quickly generate scientific decisions relying on algorithm models when facing external impacts, strengthening stability and risk resistance.

*2) Supply Chain Transaction Costs: Cost Reduction Paths through Data Governance and Intelligent Collaboration*

Relying on data governance technologies and intelligent interaction systems, the marketization of data elements fundamentally reduces supply chain transaction costs and indirectly improves resilience.

Firstly, data compliance and privacy computing technologies are used to reduce implicit costs. The marketization of data elements promotes the establishment of unified data transaction standards. Combined with privacy computing technologies such as federated learning and homomorphic encryption, data sharing of "usable but invisible" is realized on the premise of ensuring data security, which significantly reduces the negotiation costs caused by information asymmetry and the risk costs of privacy leakage. Taking the supply chain finance scenario as an example, based on zero-knowledge proof technology, financial institutions can verify the credit level of enterprises without obtaining their core data, shortening the financing review cycle from weeks to days.

Secondly, data standardization and middle platform architecture are adopted to improve collaboration efficiency. The professional data middle platform builds a unified data coding standard and classification system, integrates public and private domain data, and realizes real-time data cleaning and synchronization through ETL (Extract-Transform-Load) tools, ensuring the accuracy and timeliness of information transmission among various nodes of the supply chain. This technical architecture breaks down the internal departmental barriers of enterprises and supply chain system barriers, improving the collaboration efficiency of production, procurement, logistics and other links by more than 40%, and indirectly reducing communication costs and process redundancy costs.

*3) Digitalization of Traditional Production Factors: Multiplier Effect of Factor Empowerment and Model Integration*

Through factor digital modeling and intelligent collaboration algorithms, the marketization of data elements activates the digital multiplier effect of traditional production factors and reconstructs the operation logic of the supply chain.\

At the level of labor factors, the marketization of data elements promotes the construction of an intelligent production system of "human-machine collaboration": employee operation data is collected through computer vision technology, and the operation process is optimized combined with action recognition algorithms; relying on the personalized recommendation algorithm of the online learning platform, employees are accurately matched with skill training content, improving the adaptability of human-machine interaction. This model enhances the collaboration efficiency between labor quality and production needs, forming a positive cycle of "technological iteration - skill upgrading - productivity multiplication".

At the level of technical factors, the technology diffusion model based on knowledge graph accelerates the transformation of innovation achievements: the data element marketization platform integrates resources such as patent data and technical documents, and matches enterprises with suitable technical solutions through semantic analysis algorithms, shortening the cycle of technology introduction and implementation.

At the level of capital factors, the marketization of data elements optimizes capital allocation relying on intelligent risk control algorithms: by integrating transaction data and logistics data of supply chain enterprises, a credit evaluation model based on XGBoost (Extreme Gradient Boosting) is built to realize accurate identification of high-quality capital and efficient elimination of non-performing assets. At the same time, real-time data monitoring is used to strengthen the supervision of capital flow, maximizing the efficiency of capital utilization.

The digital transformation of traditional production factors, through the in-depth integration of algorithm models and factor resources, significantly improves the operational efficiency and flexibility of the supply chain. Based on this, this paper proposes the following hypotheses:

H2a: The marketization of data elements enhances the resilience of enterprise supply chains through the technology innovation effect driven by algorithms.

H2b: The marketization of data elements improves the resilience of enterprise supply chains through the transaction cost reduction effect of data governance and intelligent collaboration.

H2c: The marketization of data elements triggers the structural transformation of traditional production factors through factor digital modeling, thereby enhancing the resilience of enterprise supply chains.

The mechanism of how the marketization of data elements affects supply chain resilience is shown in Figure 3.

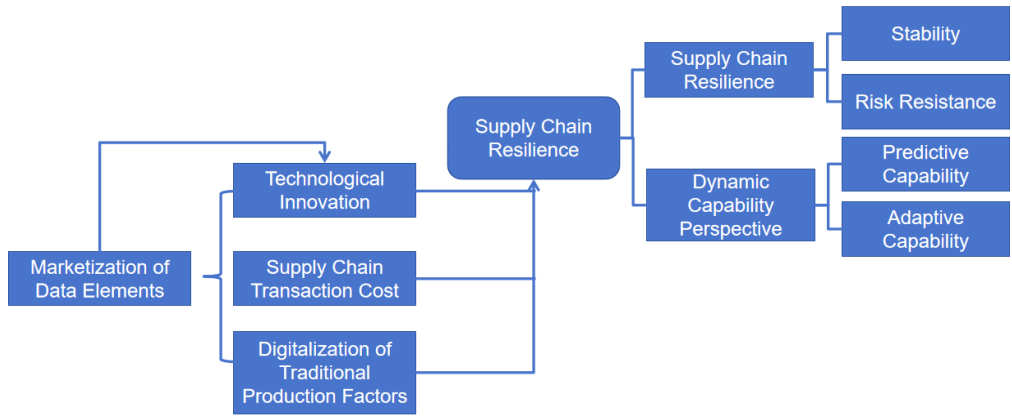

Fig. 3. Analysis framework and influence path

## III. DESIGN OF THE STUDY

### A. Model-building

To verify the causal relationship between digital element marketization and enterprise supply chain resilience, we use the Double Machine Learning (DML) model for empirical testing. Ordinary multiple linear regression has three flaws: (1) Exponential decline in training sample density with increasing covariate dimensions causes overfitting, larger parameter estimation variance, and lower quality; (2) Distance measurement fails to effectively measure dissimilarity in high-dimensional spaces; (3) Potential nonlinear relationships lead to estimation bias. Compared with traditional asymptotic DID, DML avoids the "curse of dimensionality" via cross-validation and fits this study better.

On one hand, supply chain resilience (a key indicator of high-quality enterprise development) is affected by complex macroeconomic interference. Incorporating sufficient control factors to ensure model robustness causes the "curse of dimensionality," which DML addresses by selecting high-dimensional controls via regularization, fitting effective control sets through cross-validation, and constructing confidence intervals. On the other hand, complex nonlinear variable relationships in enterprise development lead to uncorrectable bias in panel regressions (even with fixed effects), while DML captures these relationships for unbiased estimation.

Drawing on Zhou & Wang (2024), we use regional data trading platforms as proxy exogenous policy shocks for data element marketization, construct a DML model to evaluate its impact on supply chain resilience, and establish the following partially linear model:

$$SCR_{it+1} = \theta_0 Mde_{it} + g_0(X_{it}) + U_{it} \quad (1)$$

$$E[U_{it}|X_{it}, Mde_{it}] = 0 \quad (2)$$

$$Mde_{it} = m_0(X_{it}) + V_{it} \quad (3)$$

$$E[V_{it}|X_{it}] = 0 \quad (4)$$

$$X_{it} = (X_{1t}, \cdots, X_{pt}) \quad (5)$$

$SCR_{it+1}$ (dependent variable): Supply chain resilience of firm i in year t+1; $Med_{it}$ (policy variable): 1 if the region of firm i established a data trading platform in year t, otherwise 0; $\theta_0$ (key coefficient): Hypothesis H1 holds if $\theta_0 > 0$; $X_{it}$: High-dimensional control variable set; $U_{it}$, $V_{it}$: Error terms with conditional mean 0; Confounding factor $X_{it}$ affects $Mde_{it}$ via $m_0(X_{it})$ and $SCR_{it+1}$ via $g_0(X_{it})$.

If the minimum least squares estimate θ0 is obtained to get the $\hat{\theta}_0$ of Equation (6):

$$\widehat{\theta_0} = \left(\frac{1}{n}\sum_{i \in I} M\, de_{ii}^2\right)^{-1} \frac{1}{n}\sum_{i \in I} M\, de_{ii}\left(SCR_{i+1} - \widehat{g_0}(X_i)\right)(6)$$

$$\sqrt{n}(\widehat{\theta_0} - \theta_0) = \left(\frac{1}{n}\sum_{i \in I} M\, de_{ii}^2\right)^{-1} \frac{1}{\sqrt{n}}\sum_{i \in I} M\, de_{ii} U_{ii} + \left(\frac{1}{n}\sum_{i \in I} M\, de_{ii}^2\right)^{-1} \frac{1}{\sqrt{n}}\sum_{i \in I} M\, de_{ii}\left(g_0(X_{ii}) - \widehat{g_0}(X_{ii})\right)(7)$$

$$a = \left(\frac{1}{n}\sum_{i \in I} M\, de_{ii}^2\right)^{-1} \frac{1}{\sqrt{n}}\sum_{i \in I} M\, de_{ii} U_{ii}, b = \left(\frac{1}{n}\sum_{i \in I} M\, de_{ii}^2\right)^{-1} \frac{1}{\sqrt{n}}\sum_{i \in I} M\, de_{ii}\left(g_0(X_{ii}) - \widehat{g_0}(X_{ii})\right)(8)$$

Sinc $E[U_{it} \mid X_{it}, Med_{it}] = 0$, we have α ~ $N(0, \Sigma^{-1})$. However, as b → ∞, it follows that $n(\hat{\theta}_0 - \theta_0) \xrightarrow{p} \infty$. If the influence of $X_{it}$ is removed from $Med_{it}$, then $V = Med_{it} - m_0(X_{it})$ and $\hat{V} = Med_{it} - \hat{m}_0(X)$.

DML estimates the unbiased estimator $\hat{m}_0$ of $m_0$ and the unbiased estimator $\hat{g}_0$ of $g_0$ using machine learning (ML), after which the $\hat{\theta}_0$ in Equation (8) can be calculated:

$$\dot{\theta}_0 = \left(\frac{1}{n}\sum_{i \in I} \hat{V}_i\, Mde_{ii}\right)^{-1} \frac{1}{n}\dot{V}_i\left(SCR_{ii+1} - \hat{g}_0(X_{ii})\right)(8)$$

Consider the estimation bias in Equation (9):

$$\sqrt{n}(\dot{\theta}_0 - \theta_0) = (E[V_{ii}^2])^{-1}\frac{1}{\sqrt{n}}\sum_{i \in I} V_{ii}\, U_{ii} + (E[V_{ii}^2])^{-1}\frac{1}{\sqrt{n}}\sum_{i \in I}(\widehat{m_0}(X_{ii}) - m_0(X_{ii}))(\widehat{g_0}(X_{ii}) - g_0(X_{ii})) + \frac{1}{\sqrt{n}}\sum_{i \in I} V_{ii}(\widehat{g_0}(X_{ii}) - g_0(X_{ii})) \quad (9)$$

*B. Variable selection and data sources*

*1) Dependent Variable: Enterprise Supply Chain Resilience (SCR)*

Existing studies mostly measure supply chain resilience from a single dimension, which fails to fully reflect the complete chain of "risk resistance - dynamic adjustment - collaborative recovery". Combining the advantages of machine learning algorithms in fusing multi-dimensional features, this paper constructs 5 sub-indicators from three dimensions: stability, risk resistance, and dynamic adaptability, taking into account both supply chain operations and the collaborative characteristics of financial supply chains:

SCR1 (Supply Chain Cooperation Stability): Referring to the method of Lin Yue et al. (2023), it is measured by the "proportion of core partners (top 3 suppliers/customers) from the previous year that continue to cooperate in the current year", calculated as "number of continuously cooperating core partners ÷ 3". The value ranges from 0 to 1; the closer the value is to 1, the more stable the cooperation network, and the lower the credit transmission risk in the financial supply chain.

SCR2 (Supply Chain Concentration Risk): Measured by the average of "the proportion of purchases from top 3 suppliers + the proportion of sales to top 3 customers". A lower value indicates less dependence on the supply chain, stronger ability to resist single-node disruption risks, and reduced collateral value fluctuation risks in the financial supply chain.

SCR3 (Demand Forecasting Accuracy): Calculated as the reciprocal of the absolute deviation rate between the predicted value of the LSTM demand forecasting model and the actual sales volume, i.e., "1 - |predicted sales volume - actual sales volume| ÷ actual sales volume". The value ranges from 0 to 1; higher accuracy indicates stronger data-driven capabilities, which can reduce inventory financing default risks in the financial supply chain.

SCR4 (Operational Adaptation Efficiency): Measured by the logarithmic form of "working capital turnover rate × accounts receivable turnover rate", comprehensively reflecting the collaborative adaptation capability of supply chain cash flow and logistics. A higher value indicates more efficient adjustment of capital occupation structure when facing shocks, and better liquidity matching with the financial supply chain.

SCR5 (Resilience Recovery Capability): Combining the cash flow support characteristics of the financial supply chain, the calculation formula is "(net cash flow from operating activities + supply chain financing quota) ÷ current liabilities". It reflects both the enterprise's own cash flow resilience and the external support of the financial supply chain; a higher value indicates stronger recovery capability.

*2) Explanatory Variable: Marketization of Data Elements (Mde)*

Instead of the traditional dummy variable, this paper uses a continuous variable "Data Element Marketization Index" for measurement, which is more suitable for machine learning's processing needs of continuous features. The index is constructed through the following three steps:

Basic Indicator Layer: Select 3 core indicators: "transaction volume of regional data trading platforms", "number of enterprise data asset registrations", and "density of data service providers";

Weight Calculation: Use random forest regression algorithm to score the feature importance of each indicator, with weights of 0.42, 0.35, and 0.23 respectively;

Standardization: Convert raw data into an index value in the range [0,1] through Z-score standardization; a higher index indicates a deeper degree of data element marketization.

*C. Mediating Variables and Control Variables*

(1) Mediating Variables

To reveal the mechanism by which data element marketization affects resilience through the "technology - operation - finance" transmission chain, 3 mediating variables are set:

Tech_inno (Technological Innovation Capability): Measured by the standardized value of "number of enterprise patent applications (weighted score: invention patents × 2 + utility model patents × 1) + digital investment intensity (digital equipment expenditure ÷ total assets)";

Trans_cost (Supply Chain Transaction Cost): Construct a composite index by extracting keyword frequencies such as "coordination cost" and "information cost" from annual reports based on text mining algorithms, combined with "administrative expenses ÷ operating income";

Fin_sync (Financial Supply Chain Synergy Degree): Measured by the average of "supply chain financing balance ÷ total financing amount" and "core enterprise guarantee quota ÷ cooperative enterprise financing amount", reflecting the depth of synergy between supply chain and financial services.

(2) Control Variables

Combining enterprise micro-characteristics and macro environment, 11 variables are controlled, taking into account both supply chain operations and financial attributes:

| Variable Category | Variable Name | Definition and Calculation Method |
| --- | --- | --- |
| Enterprise Basic Characteristics | Size | Natural logarithm of total assets at the end of the period |
| | Lev | Total liabilities ÷ total assets (reflecting the constraint of financial leverage on supply chain resilience) |
| | Roa | Net profit ÷ average total assets (measuring the supporting capacity of profitability for resilience) |
| Supply Chain Operational Characteristics | Inv_turn | Operating cost ÷ average inventory (inventory turnover efficiency) |
| | Fix_ratio | Net fixed assets ÷ total assets (reflecting the impact of production rigidity on resilience) |
| Corporate Governance Characteristics | Board_size | Natural logarithm of the number of board members |
| | Dual | Dummy variable: 1 if chairman and CEO are the same person, 0 otherwise |
| | Top1 | Shareholding ratio of the largest shareholder (reflecting the impact of equity concentration on decision-making efficiency) |
| Financial Attribute Characteristics | Cash_ratio | Monetary funds ÷ current liabilities (short-term liquidity reserves) |
| | Soe | Dummy variable: 1 if state-owned enterprise, 0 otherwise |
| Growth Characteristics | Growth | (Current year operating income - previous year operating income) ÷ previous year operating income (reflecting the impact of growth pressure on resilience) |

## IV. ANALYSIS OF EMPIRICAL RESULTS

### A. Data Element Marketization's Policy Effect on Enterprise Supply Chain Resilience

Data element marketization transforms data resources into data elements, and continuous data input realizes its production value. On one hand, it fosters a sound environment for data processing and governance, generating standardized, high-quality, scalable data products that become enterprise data assets. Applied in demand forecasting, supplier stability analysis, and financial risk identification, these assets enhance supply chain responsiveness and resilience. On the other hand, it spurs data-enabled platforms that break inter-enterprise boundaries via data innovation, promoting industrial and supply chain data sharing to form a collaborative network, strengthening supply chain resilience.

This study uses Double Machine Learning (random forest algorithm, 5-fold cross-validation, with year and firm fixed effects) to verify the causal effect. Benchmark regression results (Table 1) show data element marketization significantly boosts supply chain resilience across dimensions: 10% significance for stability (SCR1) and 1% significance for risk resistance (SCR2), predictive capability (SCR3), adaptation capability (SCR4), and recovery capability (SCR5). Hypothesis H1 is validated.

TABLE I. BENCHMARK REGRESSION RESULTS

|  | (1) SCR1 | (2) SCR2 | (3) SCR3 | (4) SCR4 | (5) SCR5 |
|---|---|---|---|---|---|
| Mde | 0.038* | 0.081*** | 0.012*** | 0.010*** | 0.123*** |
|  | (1.89) | (5.09) | (5.48) | (13.62) | (9.17) |
| Constant | -0.086*** | -0.082** | 0.006*** | 0.002 | -0.147 |
|  | (-5.02) | (-2.13) | (7.59) | (0.61) | (-1.06) |
| Control Variables | Yes | Yes | Yes | Yes | Yes |
| Year Fixed Effects | Yes | Yes | Yes | Yes | Yes |
| Firm Fixed Effects | Yes | Yes | Yes | Yes | Yes |
| N | 39,521 | 41,830 | 44,435 | 40,968 | 44,502 |
| Mde | 0.038* | 0.081*** | 0.012*** | 0.010*** | 0.123*** |

*Notes: ***, *, * indicate significance at 1%, 5%, 10% levels; t-statistics in parentheses (same below).

## B. Robustness Tests

Exclude firms in municipalities (due to special policy advantages) and regions with inactive data trading platforms. Results (Table 2) remain significant.

TABLE II. SAMPLE EXCLUSION RESULTS

|  | (1) SCR1 | (2) SCR2 | (3) SCR3 | (4) SCR4 | (5) SCR5 |
|---|---|---|---|---|---|
| Mde | 0.012*** | 0.074** | 0.050*** | 0.008*** | 0.158*** |
|  | (5.65) | (2.18) | (5.17) | (13.60) | (4.26) |
| Constant | -0.193** | -0.025*** | 0.006*** | -0.078*** | -0.150** |
|  | (-5.13) | (-2.62) | (7.57) | (3.39) | (-1.93) |
| Controls/Year/Firm FE | Yes/Yes/Yes | Yes/Yes/Yes | Yes/Yes/Yes | Yes/Yes/Yes | Yes/Yes/Yes |
| N | 29,445 | 31,160 | 33,100 | 30,518 | 33,195 |

Add a dummy variable for the "Broadband China" pilot policy (confounding factor for data infrastructure). Results (Table 3) stay robust.

TABLE III. POLICY INTERFERENCE EXCLUSION

|  | (1) SCR1 | (2) SCR2 | (3) SCR3 | (4) SCR4 | (5) SCR5 |
|---|---|---|---|---|---|
| Mde | 0.018*** | 0.080*** | 0.039*** | 0.006*** | 0.100*** |
|  | (3.96) | (3.53) | (6.63) | (2.75) | (3.30) |
| Constant | -0.137 | -0.028 | 0.001** | -0.129 | -0.122 |
|  | (-1.07) | (-0.93) | (2.25) | (1.42) | (-0.91) |
| N | 39,548 | 44,212 | 44,540 | 41,448 | 44,295 |

Use neural network instead of random forest. Results (Table 4) confirm significance.

TABLE IV. Neural Network Algorithm Results

|  | (1) SCR1 | (2) SCR2 | (3) SCR3 | (4) SCR4 | (5) SCR5 |
|---|---|---|---|---|---|
| Mde | 0.016*** | 0.090*** | 0.043*** | 0.011** | 0.153** |
|  | (3.18) | (8.57) | (4.33) | (2.15) | (2.46) |
| Constant | -0.141 | -0.025*** | 0.020*** | -0.217*** | -0.183*** |
|  | (-1.39) | (-10.23) | (8.87) | (5.41) | (-3.50) |
| Controls/FE | Yes/Yes | Yes/Yes | Yes/Yes | Yes/Yes | Yes/Yes |
| N | 39,521 | 41,830 | 44,435 | 40,968 | 44,502 |

Adjust sample split ratio from 1:5 to 1:4 (4-fold cross-validation). Results (Table 5) remain consistent.

TABLE V. Adjusted Sample Split Results

|  | (1) SCR1 | (2) SCR2 | (3) SCR3 | (4) SCR4 | (5) SCR5 |
|---|---|---|---|---|---|
| Mde | 0.035** | 0.088*** | 0.130** | 0.034 | 0.086** |
|  | (6.15) | (7.55) | (2.05) | (1.88) | (2.46) |
| Constant | -0.211** | -0.193** | 0.036*** | -0.190** | 0.288*** |
|  | (-2.94) | (-2.14) | (2.69) | (-2.12) | (3.85) |
| Controls/FE | Yes/Yes | Yes/Yes | Yes/Yes | Yes/Yes | Yes/Yes |
| N | 39,521 | 41,830 | 44,435 | 40,968 | 44,502 |

V. Conclusion

This study integrates economic theories (e.g., Transaction Cost Theory) with machine learning to explore the synergistic optimization mechanism of SCM and FSCM, focusing on verifying how data element marketization enhances supply chain resilience through technical empowerment. Empirically, using the DML model with 5-fold cross-validation, benchmark regression shows data element marketization (Mde) significantly improves all five dimensions of supply chain resilience—boosting cooperation stability (SCR1) at the 10% significance level and strengthening risk resistance (SCR2), demand forecasting accuracy (SCR3), operational adaptation efficiency (SCR4), and recovery capability (SCR5) at the 1% significance level—by breaking "data silos" via distributed data sharing, algorithm collaboration, and trusted data interaction. The multi-channel transmission mechanism also validates Hypotheses H2a-H2c: the technological innovation effect (algorithm penetration like GAN and optimization via open API), transaction cost reduction effect (unified data standards and privacy computing cutting costs by over 40%), and factor digitalization effect (digital modeling of labor, technology, and capital activating traditional factor multipliers) all drive resilience enhancement. Robustness tests (sample exclusion, policy interference exclusion, neural network algorithm replacement, sample split adjustment) further confirm the core conclusion's reliability by eliminating confounding factors. The study contributes theoretically by constructing a "data-algorithm-resilience" framework integrating economic theories and machine learning, methodologically by using DML to address traditional regression biases, and practically by providing insights—enterprises can leverage transfer learning and federated learning for digital transformation, while policymakers should prioritize data transaction standards and intelligent service platform construction. Its limitations include a focus on manufacturing (restricting generalizability), and future research could expand samples to service/agricultural supply chains, explore heterogeneous effects across enterprise sizes/regions, and integrate digital twins with supply chain financial services.